\documentclass[10pt,twocolumn,letterpaper]{article}

\usepackage{cvpr}              

\definecolor{cvprblue}{rgb}{0.21,0.49,0.74}
\usepackage[pagebackref,breaklinks,colorlinks,allcolors=cvprblue]{hyperref}
\usepackage{graphicx}
\usepackage{comment}
\usepackage{tabularx}
\usepackage{booktabs}
\usepackage{multirow}
\usepackage{float}
\usepackage{amssymb}
\usepackage{soul}
\usepackage{xcolor}
\usepackage{color, colortbl}
\usepackage{comment}

\definecolor{Gray}{gray}{0.1}
\definecolor{LightCyan}{rgb}{0.88,1,1}
\definecolor{LightYellow}{rgb}{1,1,0.88}
\definecolor{LightPurple}{rgb}{1,0.88,1}
\definecolor{LightRed}{rgb}{1,0.88,0.88}
\definecolor{LightGreen}{rgb}{0.88,1,0.88}
\definecolor{LightBlue}{rgb}{0.88,0.88,1}

\def\paperID{8538} 
\def\confName{CVPR}
\def\confYear{2025}

\title{Video-Panda: Parameter-efficient Alignment for Encoder-free Video-Language Models}

\author{%
    Jinhui Yi*$^{1,2}$ \quad
    Syed Talal Wasim*$^{1,2}$ \quad
    Yanan Luo*$^{1}$ \quad
    Muzammal Naseer$^{3}$ \quad
    Juergen Gall$^{1,2}$ \\
    \vspace{-3mm} \\ 
    $^{1}$University of Bonn \quad
    $^{2}$ Lamarr Institute for Machine Learning and Artificial Intelligence \\
    $^{3}$Khalifa University 
    * Equal Contribution
}

\begin{document} 
\maketitle

\begin{abstract}
We present an efficient encoder-free approach for video-language understanding that achieves competitive performance while significantly reducing computational overhead. Current video-language models typically rely on heavyweight image encoders (300M-1.1B parameters) or video encoders (1B-1.4B parameters), creating a substantial computational burden when processing multi-frame videos. Our method introduces a novel Spatio-Temporal Alignment Block (STAB) that directly processes video inputs without requiring pre-trained encoders while using only 45M parameters for visual processing - at least a 6.5$\times$ reduction compared to traditional approaches. The STAB architecture combines Local Spatio-Temporal Encoding for fine-grained feature extraction, efficient spatial downsampling through learned attention and separate mechanisms for modeling frame-level and video-level relationships. Our model achieves comparable or superior performance to encoder-based approaches for open-ended video question answering on standard benchmarks. The fine-grained video question-answering evaluation demonstrates our model's effectiveness, outperforming the encoder-based approaches Video-ChatGPT and Video-LLaVA in key aspects like correctness and temporal understanding. Extensive ablation studies validate our architectural choices and demonstrate the effectiveness of our spatio-temporal modeling approach while achieving 3-4$\times$ faster processing speeds than previous methods. Code is available at \url{https://jh-yi.github.io/Video-Panda}.
\end{abstract}    
\section{Introduction}
\label{sec:intro}

Recently, large language models (LLMs) have demonstrated remarkable capabilities in understanding and generating text, catalyzing significant advancements in vision-language modeling~\cite{zhao2023llmsurvey, touvron2023llama, liu2023llava, achiam2023gpt4}. While initial efforts focused on image understanding, recent works have extended these capabilities to video comprehension, enabling more complex spatio-temporal reasoning~\cite{maaz2023videochatgpt,li2023videochat,zhang2023videollama}. However, current video-language models face two major challenges.

First, existing video-language models typically rely on either heavyweight image encoders (300M to 1.1B parameters)~\cite{zhang2024llamaadapter, jin2024chatunivi, maaz2023videochatgpt, lin2023videollava} or even larger video encoders (1B to 1.4B parameters)~\cite{li2023videochat, zhang2023videollama, wang2024internvideo2}. Some approaches even combine both types of encoders for enhanced feature extraction~\cite{maaz2024videogpt+}. These large encoders create substantial computational overhead, particularly when processing videos with multiple frames that require repeated passes through the encoder. Furthermore, the alignment between these encoders and language models is typically achieved through either simple linear projections~\cite{zhang2024llamaadapter}, which ignore spatio-temporal relations, or more complex mechanisms that combine linear projections with additional resampling modules such as query-transformers~\cite{li2023videochat, zhang2023videollama}, adding further computational complexity.

\begin{figure}[t]
  \centering
    \includegraphics[width=\linewidth]{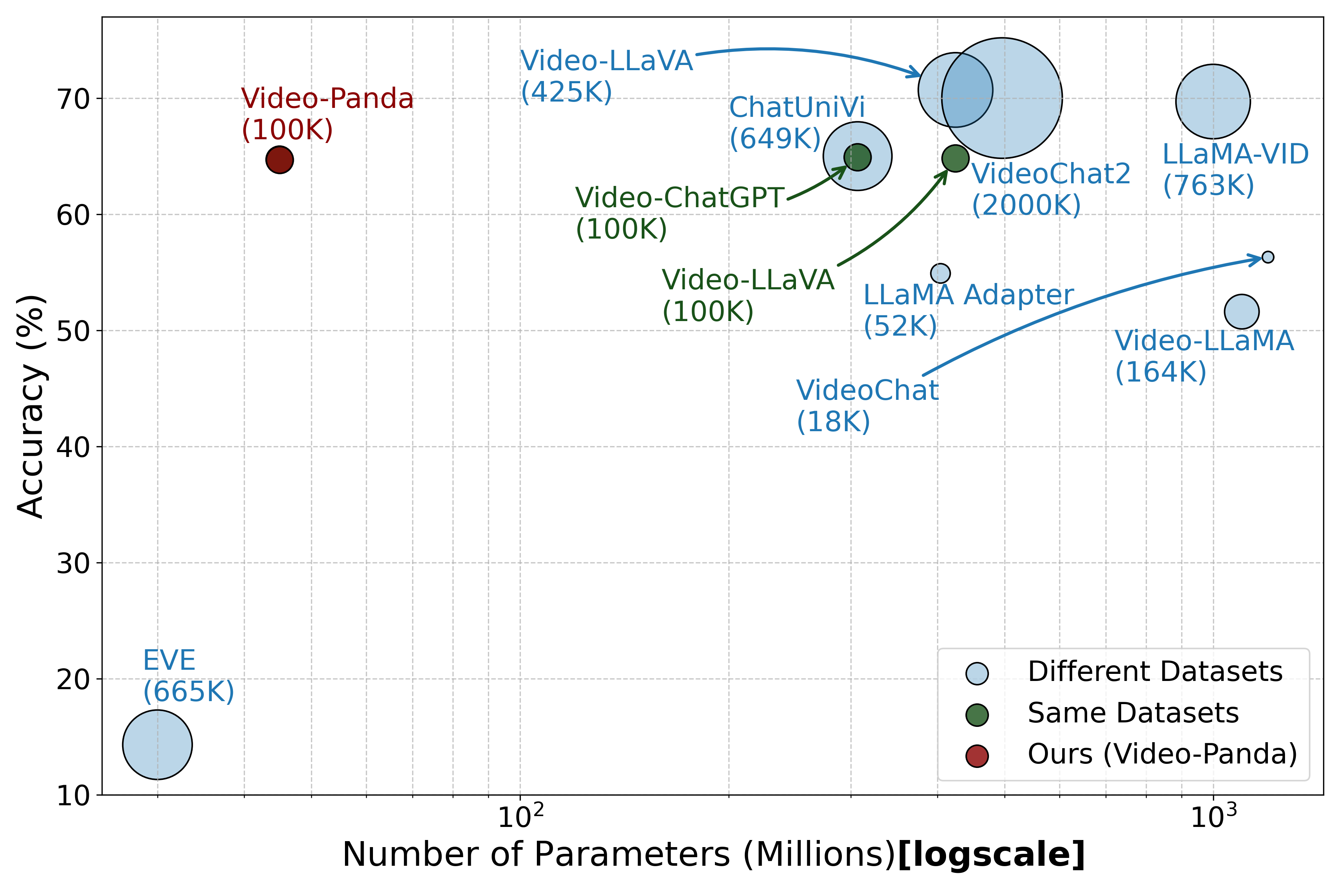} 
   \caption{Model performance on MSVD-QA versus the model size of the visual component in logarithmic scale. The bubble size indicates the amount of finetuning data (in thousands). Models using the same training dataset as ours (100K samples) are shown in dark green, while those using different datasets are in blue.}
   \label{fig:teaser}
\end{figure}

Second, current approaches often struggle with the inherent complexity of video understanding. Recent work has shown that naive adaptations of image-language architectures to video comprehension result in significant performance degradation when trained solely on video data~\cite{li2023videochat}. This suggests that effective video-language models require specialized architectures that can capture the unique spatio-temporal relationships present in video content, rather than treating videos as simple sequences of images.

Given these considerations, we ask: \textit{Is it possible to develop an efficient video-language model that bypasses the need for heavyweight encoders while maintaining state-of-the-art accuracy?} To address this question, we introduce a novel encoder-free video-language model that achieves remarkable performance with significantly reduced computational overhead. The key to our approach is a specialized spatio-temporal alignment block that effectively bridges visual and linguistic information without requiring image or video encoders. 

As shown in \autoref{fig:teaser}, our approach achieves strong performance on MSVD-QA~\cite{chen2011msvd} video-language benchmarks but requires only a fraction of the parameters typically dedicated to visual encoding in traditional approaches. Our key contributions include:
\begin{itemize}
    \item A novel encoder-free video-language model that achieves strong open-ended video question answering performance while using only 45M parameters for visual processing - a $6.5\times$ reduction compared to Video-ChatGPT (307M) and $9 \times$ reduction compared to Video-LLaVA (425M).
    \item Strong performance on fine-grained video question answering metrics, outperforming encoder-based approaches on key aspects like correctness and temporal understanding.
    \item A specialized spatio-temporal alignment block that processes videos 3-4$\times$ faster than the encoder-based approaches Video-ChatGPT or Video-LLaVA while maintaining competitive accuracy.
\end{itemize}
\section{Related Work}
\label{sec:related_work}

\subsection{Large Language Models}
The landscape of Large Language Models (LLMs)~\cite{zhao2023llmsurvey, ouyang2022instructgpt, brown2020gpt3, achiam2023gpt4, bubeck2023sparks, touvron2023llama, touvron2023llama2} has evolved significantly with the introduction of various commercial and open-source models. Following the release of ChatGPT~\cite{openai2022chatgpt}, the AI community responded with open-source alternatives like LLaMA~\cite{touvron2023llama}, Vicuna~\cite{vicuna2023}, Alpaca~\cite{taori2023alpaca}, and LLaMA 2~\cite{touvron2023llama2}. These models underwent instruction tuning to facilitate human-AI conversations. Notable developments include InstructGPT~\cite{ouyang2022instructgpt}, which was trained based on GPT-3~\cite{brown2020gpt3} with 175 billion parameters through human preference alignment. While these models demonstrated powerful reasoning capabilities, they were initially limited to text-only interactions, prompting research into multimodal extensions.

\subsection{Image-Language Models}
The integration of Large Language Models (LLMs) with Large Vision Models (LVMs) has produced two distinct approaches to building Image-Language Models. The first and more established approach uses encoder-based architectures, as seen in commercial offerings like Claude-3V~\cite{anthropic2024claude3v}, Grok-1.5V~\cite{xai2024grok}, MM1~\cite{mckinzie2024mm1}, Gemini series~\cite{team2023gemini}, and GPT-4V~\cite{openai2023gpt4v}.  In the open-source domain, models like BLIP-2~\cite{li2023blip2}, MiniGPT-4~\cite{zhu2024minigpt4}, and LLaVA~\cite{liu2023llava, liu2023llava1.5} employed learnable modules to project image features into the language space.

More recently, an encoder-free approach has emerged, pioneered by Fuyu-8B~\cite{bavishi2023fuyu-8b}, which processes image inputs directly through a decoder-only network. This architecture naturally handles high-resolution images with arbitrary aspect ratios, avoiding the constraints of traditional encoder-based systems. Recent advances in this direction include EVE~\cite{diao2024eve}, which introduces a novel training recipe focusing on a unified decoder representation and enhanced visual recognition capabilities. These encoder-free models effectively address several key challenges: resolution constraints, deployment overhead, and capacity matching between vision and language components. Using minimal public training data, models like EVE~\cite{diao2024eve} have demonstrated competitive performance with encoder-based approaches while requiring fewer computational resources.

\begin{figure*}[t]
  \centering
    \includegraphics[width=\linewidth]{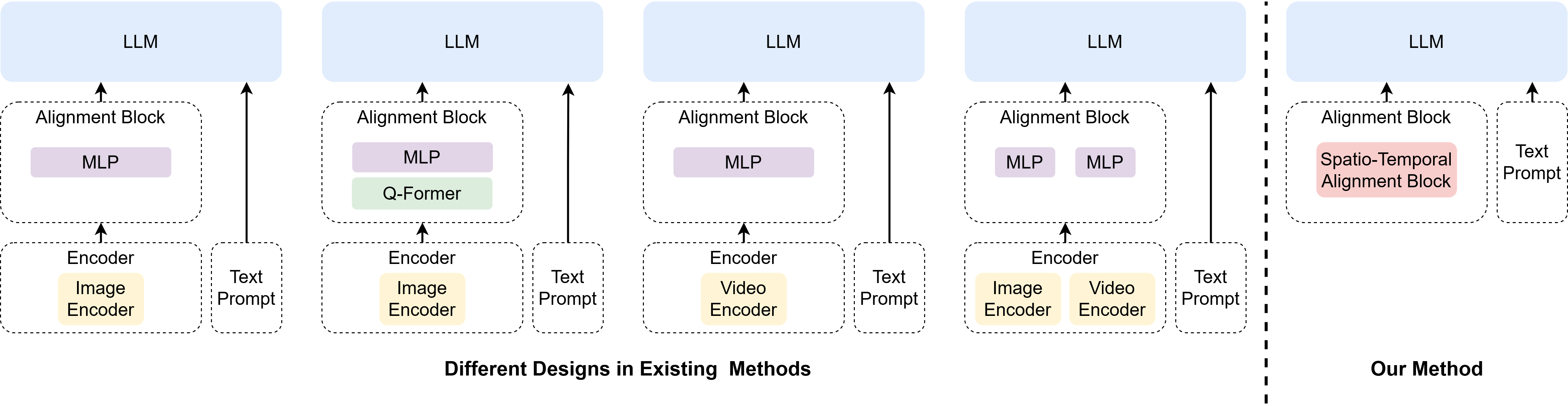} 
    \vspace{-5mm}
    \caption{\textbf{Existing video-language model architectures:} From left to right: Early approaches use image encoders for both image and video inputs. The alignment module aligns the embeddings of the visual modality with the language modality. The integration of Q-Former improved this alignment. Instead of a single encoder, dual encoder approaches have separate encoders for images and videos where the alignment block consists of projection layers. The additional encoders, however, make these models very heavy where the alignment module and encoders have at least 300M and sometimes over 1B parameters. In contrast, our encoder-free design (rightmost) directly processes video inputs through a novel spatio-temporal alignment block (STAB). It eliminates the need for heavyweight pretrained encoders and requires less than 50M parameters.}
   \label{fig:structure_comparison}
\end{figure*}

\subsection{Video-Language Models}

When extending language models to video understanding, several approaches have been proposed to handle the additional temporal dimension. Early video-capable models like Video-ChatGPT~\cite{maaz2023videochatgpt} introduced temporal encoding components but faced challenges regarding computational cost and limitations of the context window.

VideoChat~\cite{li2023videochat} initiated an end-to-end approach by integrating video foundation models with LLMs through a learnable neural interface. This system excelled in spatiotemporal reasoning and event localization, demonstrating the potential of chat-centric video understanding. Building upon this, VideoChat2~\cite{li2024mvbench} further advanced temporal understanding capabilities through progressive multi-modal training with diverse instruction-tuning data. LLaMA-Adapter V2~\cite{gao2023llamaadapterv2} proposed an efficient approach by unlocking more learnable parameters and introducing an early fusion strategy for visual tokens. Their joint training paradigm effectively balanced image-text alignment and instruction following tasks. Similarly, LLaMA-VID~\cite{li2025llamavid} used a dual-token strategy - using context and content tokens for each frame, enabling the processing of hour-long videos while preserving critical information.

Video-LLaMA~\cite{zhang2023videollama} addressed multimodal integration by incorporating both visual and auditory content through a Video Q-former for temporal understanding and an Audio Q-former built on ImageBind for audio processing. Chat-UniVi~\cite{jin2024chatunivi} took a different approach by utilizing dynamic visual tokens to create a unified representation for both images and videos, employing a multi-scale representation to capture both high-level concepts and low-level details. ST-LLM~\cite{liu2024stllm}, proposed as dynamic masking strategy which improved both the efficiency and robustness of the video VLM. Video-LLaVA~\cite{lin2023videollava} introduced a unified approach that aligns visual representations before projecting them into the language space. Combined with joint training on mixed image and video datasets, this method demonstrated strong performance across modalities. 
In this work, we address a notable gap in the literature and present the first encoder-free video-language model.

\section{Method}
\label{sec:methodology}

Recent advances in video-language understanding have been primarily driven by extending Large Language Models (LLMs) through various architectural paradigms, as illustrated in \autoref{fig:structure_comparison}. Traditional approaches rely heavily on pretrained vision encoders (300M-1.4B parameters) to extract frame-level features, either adapting image encoders for video processing or employing specialized video encoders. These designs typically incorporate additional alignment modules like Q-Former~\cite{li2023blip2} or MLPs to bridge the visual-linguistic gap. While effective, such approaches inherit significant computational overhead from their reliance on heavyweight encoders and complex alignment strategies. In contrast, we propose a radically different design philosophy: an encoder-free architecture that directly processes video inputs through a specialized spatio-temporal alignment block (STAB). This approach not only eliminates the dependency on pretrained vision encoders but also enables explicit modeling of video dynamics through dedicated spatio-temporal mechanisms. Recent works like Video-ChatGPT~\cite{maaz2023videochatgpt} and Video-LLaVA~\cite{lin2023videollava} have shown strong performance using encoder-based approaches, with some methods further incorporating Q-former modules adapted from BLIP-2~\cite{li2023blip2} to improve temporal modeling. However, these approaches still fundamentally rely on frame-level features extracted by vision encoders that lack spatio-temporal modeling capabilities.

The aforementioned observations lead us to develop an encoder-free architecture that directly processes video inputs through a specialized spatio-temporal alignment mechanism. In the following sections, we first present our Spatio-Temporal Alignment Block (STAB) (\autoref{subsec:model_architecture_stab}), followed by our training procedure (\autoref{subsec:training_procedure}). Our approach achieves competitive performance with significantly reduced parameters.

\begin{figure*}[t]
  \centering
    \includegraphics[width=\linewidth]{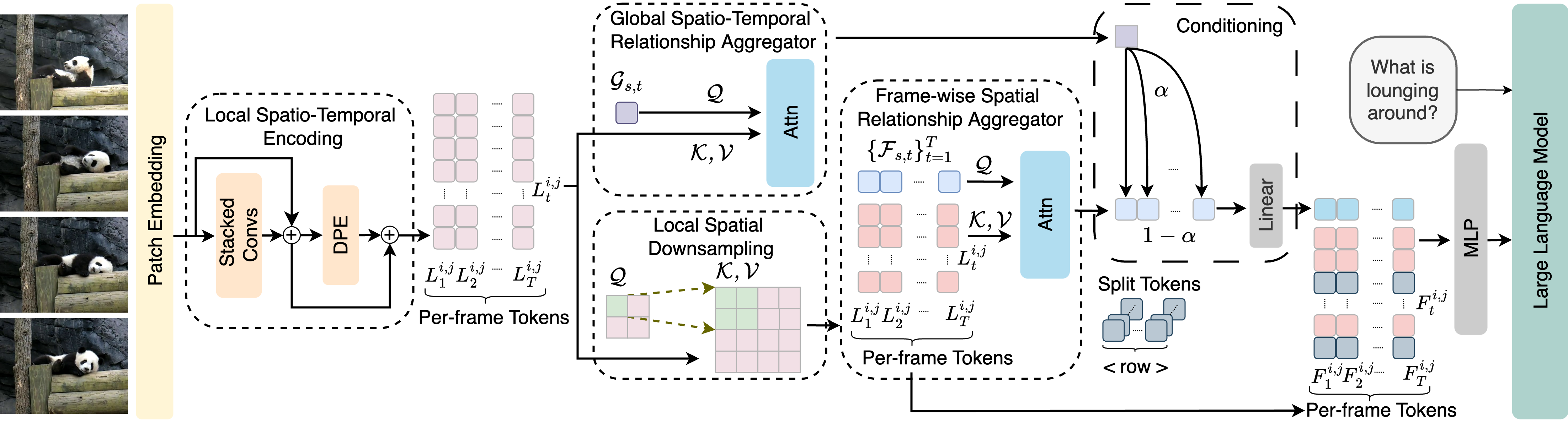} 
    \vspace{-5mm}
    \caption{\textbf{Detailed architecture of our Spatio-Temporal Alignment Block (STAB):} The input video is first converted into patches. The Local Spatio-Temporal Encoding (LSTE) uses 3D convolutions to model spatio-temporal relations and adds a 3D convolution dynamic position encoding (DPE) to encode position with respect to the local spatio-temporal window. As a result, we obtain per-frame tokens with positional encoding. The tokens are then processed in two ways. While the Global Spatio-Temporal Relationship Aggregator (GSTRA) at the top captures video-level context, the Frame-wise Spatial Relationship Aggregator (FSRA) at the bottom captures spatial context within each frame. To reduce the cost, we perform a Local Spatial Downsampling (LSD) to reduce the spatial dimension for each token. The video-level context tokens and the frame-wise spatial tokens are then linearly combined through learnable weighted fusion ($\alpha$), producing a frame-specific context token. These context tokens are then prepended to their respective frame's flattened spatial tokens, with $\texttt{<row>}$ split tokens inserted to demarcate row boundaries in the spatial layout. This combination of global context and preserved spatial structure enables effective video understanding while maintaining computational efficiency.}
   \label{fig:framework}
\end{figure*}

\subsection{Spatio-Temporal Alignment Block (STAB)}
\label{subsec:model_architecture_stab}

Our Spatio-Temporal Alignment Block (STAB) is illustrated in \autoref{fig:framework}. Since we do not have a pre-trained video encoder where the features need to be aligned with the embedding of the LLM, STAB aligns the video directly with the LLM. A core aspect is the modeling and handling of spatial and temporal information in a video. First, the Local Spatio-Temporal Encoding (LSTE) encodes local information within a small spatio-temporal window. We then separate global context at the video level, which is captured by the Global Spatio-Temporal Relationship Aggregator (GSTRA), and more fine-grained context at the frame level, which is captured by the Frame-wise Spatial Relationship Aggregator (FSRA). While GSTRA models spatio-temporal relationships across the entire video, FSRA models spatial relationships across each frame. In our ablation study, we demonstrate that both contain complementary information and their combination improves the results for open-ended video question answering, which includes questions related to the overall video content as well as questions that require a deeper understanding of the temporal information in the video. The corresponding tokens are then fused and aligned with the LLM embedding. We will now discuss each step in detail.

\noindent\textbf{Patch Embedding:} We first divide each frame of a video $\mathbf{V} \in \mathbb{R}^{T \times H \times W \times 3}$ with $T$ frames and spatial resolution $H\times W$ into non-overlapping patches:
\begin{equation}
    \mathbf{P} = f_{\text{patch}}(\mathbf{V}) \in \mathbb{R}^{T \times h \times w \times d}
\end{equation}
where $h = H/p$, $w = W/p$ for patch size $p$, and $d$ is the embedding dimension.

\noindent\textbf{Local Spatio-Temporal Encoding (LSTE):} Videos exhibit rich temporal dynamics that evolve through local spatial neighborhoods. To capture these patterns efficiently, we design a local spatio-temporal aggregation mechanism that processes information within small spatio-temporal windows, allowing each token to aggregate context from its immediate spatial and temporal neighborhood while maintaining computational efficiency:
\begin{align}
    \mathbf{L}_{st}' &= \text{Conv3D}_3(\text{Conv3D}_2(\text{Conv3D}_1(\mathbf{P}))) + \mathbf{P} \\
    \mathbf{L}_{st} &= \mathbf{L}_{st}' + \text{DPE}(\mathbf{L}_{st}')
\end{align}
where $\text{Conv3D}_1$ and $\text{Conv3D}_3$ use kernel size $(1\times1\times1)$ for channel reduction and restoration, $\text{Conv3D}_2$ uses kernel size $(3\times1\times1)$ for temporal context, and $\text{DPE}$ is a dynamic positional encoder implemented~\cite{li2022uniformer} as depthwise 3D convolution that uses kernel size $(3\times3\times3)$ for joint spatio-temporal context.

\noindent\textbf{Local Spatial Downsampling (LSD):} While maintaining fine-grained temporal resolution is crucial for video understanding, the spatial resolution can be efficiently compressed without significant loss of information. We achieve this through a window-based spatial aggregation mechanism that both enriches local spatial context and reduces computational overhead through adaptive downsampling. First, we define the scaled dot-product attention with learnable projection matrices for query, key, and value inputs:
\begin{equation}
    \text{Attn}(\mathcal{Q}, \mathcal{K}, \mathcal{V}) = \text{softmax}\left(\frac{P_q(\mathcal{Q})P_k(\mathcal{K})^T}{\sqrt{d}}\right)P_v(\mathcal{V})
\end{equation}
where $P_q$, $P_k$, and $P_v$ are learnable projection functions for query, key, and value transformations respectively. Then, for each window at position $(i,j)$ in frame $t$
\begin{align}
    \mathbf{W}_{i,j} &\in \mathbb{R}^{1 \times d} \text{ (learnable query)}\\ 
    \mathbf{K}_{i,j,t} &= \mathbf{L}_{st,t,i:i+2,j:j+2} \in \mathbb{R}^{4 \times d} \text{ (2$\times$2 window)}\\ 
    \mathbf{L}_{st,i,j,t}^{d} &= \text{Attn}(\mathbf{W}_{i,j}, \mathbf{K}_{i,j,t}, \mathbf{K}_{i,j,t})
\end{align}
yields the downsampled representation $\mathbf{L}_{st}^{d} \in \mathbb{R}^{T \times h' \times w' \times d}$ where $h' = h/2$ and $w' = w/2$.

\noindent\textbf{Frame-wise Spatial Relationship Aggregator (FSRA):} Different frames in a video can vary significantly in their spatial content and importance. To capture this frame-specific information effectively, we design a global spatial aggregator that summarizes each frame's content into a compact representation while preserving its unique spatial characteristics:
\begin{align}
    \mathbf{W}_t &\in \mathbb{R}^{1 \times d} \text{ (frame-specific query)}\\
    \mathbf{K}_t &= \text{Flatten}(\mathbf{L}^d_{st,t}) \in \mathbb{R}^{(h'w') \times d} \text{ (frame tokens)}\\
    \mathbf{F}_{s,t} &= \text{Attn}(\mathbf{W}_t, \mathbf{K}_t, \mathbf{K}_t)\,,
\end{align}
producing frame summaries $\mathbf{F}_s = \{\mathbf{F}_{s,t}\}_{t=1}^T \in \mathbb{R}^{T \times d}$.

\noindent\textbf{Global Spatio-Temporal Relationship Aggregator (GSTRA):} Understanding a video often requires capturing both frame-level details and video-wide context. We achieve this through a two-step process: first aggregating global video context, then using this to condition frame-level representations, allowing each frame to maintain its own characteristics while being informed by the broader video context \cite{li2023uniformerv2, wasim2023vita}:
\begin{align}
    \mathbf{W}_{st} &\in \mathbb{R}^{1 \times d} \text{ (global spatio-temporal query)}\\
    \mathbf{K}_{st} &= \text{Flatten}(\mathbf{L}_{st}) \in \mathbb{R}^{(Th'w') \times d} \text{ (all tokens)}\\
    \mathbf{G}_{st} &= \text{Attn}(\mathbf{W}_{st}, \mathbf{K}_{st}, \mathbf{K}_{st}) \in \mathbb{R}^{1 \times d}\,.
\end{align}
Then, we condition each frame summary with the global context:
\begin{equation}
    \mathbf{F}_{r,t} = f_{\text{proj}}(\alpha \mathbf{F}_{s,t} + (1-\alpha)\mathbf{G}_{st})
\end{equation}
where $\alpha \in \mathbb{R}^d$ is learned and $f_{\text{proj}}$ is a shared linear projection, yielding the final aggregated representation $\mathbf{F}_r \in \mathbb{R}^{T \times d}$.

\noindent\textbf{Final Token Sequence Construction:} To maintain the structural information of the video while providing both local and global context to the language model, we carefully construct the final token sequence. First, we organize the downsampled tokens $\mathbf{L}^d_{st}$ into frame-wise sequences with row delimiter tokens:
\begin{equation}
    \mathbf{F}_{\text{tokens}} = \{\text{Add-Row-Split}(\mathbf{L}^d_{st,t})\}_{t=1}^T\,.
\end{equation}
Here, we insert row-split tokens $\texttt{<row>}$ after each row of spatial tokens in a frame, helping maintain 2D spatial structure. Then, we combine these with the conditioned frame representations:
\begin{equation}
    \mathbf{F} = \{[\mathbf{F}_{r,t}; \mathbf{F}_{\text{tokens},t}]\}_{t=1}^T
\end{equation}
placing each frame's conditioned representation $\mathbf{F}_{r,t}$ at the start of its corresponding token sequence. The final visual tokens are obtained by flattening $\mathbf{F}$ and projecting it through an MLP to the language model's embedding space:
\begin{equation}
\mathbf{V}_{\text{tokens}} = \text{MLP}(\text{Flatten}(\mathbf{F}))
\end{equation}
while maintaining spatial structure through the split tokens.

\subsection{Training Procedure}
\label{subsec:training_procedure}
Our training strategy follows three carefully designed stages to ensure stable optimization and effective knowledge transfer. First, we define two core losses used throughout training. For visual alignment, we use a distillation loss that aligns our predicted visual tokens with a teacher model. Let $\mathbf{O} = (o_1, ..., o_N) \in \mathbb{R}^{N \times d}$ be the output tokens from a video-language model, where $N$ is the sequence length and $d$ is the embedding dimension. We can partition these tokens into visual and non-visual tokens:
\begin{equation}
    \mathbf{O} = [\mathbf{V}_{\text{pred}}; \mathbf{O}_{\text{other}}]
\end{equation}
where $\mathbf{V}_{\text{pred}} \in \mathbb{R}^{M \times d}$ represents the $M$ predicted visual tokens and $\mathbf{O}_{\text{other}}$ contains the remaining textual and special tokens. Given an input video, let $\mathbf{V}_{\text{teacher}} \in \mathbb{R}^{M \times d}$ be the visual tokens extracted from a pre-trained teacher model. The individual predicted tokens and teacher tokens are denoted as $\mathbf{v}_{\text{pred},i}$ and $\mathbf{v}_{\text{teacher},i}$, respectively. The distillation loss is then defined as:
\begin{equation}
    \mathcal{L}_{\text{distill}} = -\frac{1}{M}\sum_{i=1}^M \frac{\mathbf{v}_{\text{pred},i}^\top \mathbf{v}_{\text{teacher},i}}{\|\mathbf{v}_{\text{pred},i}\| \|\mathbf{v}_{\text{teacher},i}\|}\,.
\end{equation}

We use LanguageBind~\cite{zhu2023languagebind} as our teacher model for extracting $\mathbf{V}_{\text{teacher}}$. For text generation, we use a standard cross-entropy loss over the textual tokens in $\mathbf{O}_{\text{other}}$:
\begin{equation}
    \mathcal{L}_{\text{text}} = -\sum_{t \in \mathcal{T}} \log p(o_t|o_{<t}, \mathbf{O})
\end{equation}
where $\mathcal{T}$ denotes the indices of textual tokens in $\mathbf{O}_{\text{other}}$, and $p(o_t|o_{<t}, \mathbf{O})$ is the model's prediction probability for token $o_t$ given previous tokens.

\noindent\textbf{Initial Alignment Stage:} The primary goal of this stage is to establish a foundational understanding of video content through our spatio-temporal alignment mechanism. Using 351k video-text pairs from WebVid~\cite{bain2021frozen} (half of the full dataset), we train only the STAB components while keeping the LLM frozen. This stage is crucial for preventing training collapse and ensuring stable convergence, as it allows the alignment components to develop basic video understanding capabilities without interfering with the LLM's pre-trained knowledge.

\noindent\textbf{Visual-Language Integration Stage:} Using the same WebVid dataset, we fine-tune the entire model end-to-end to develop joint visual-language understanding. This stage allows the LLM to adapt its language understanding capabilities to video-specific contexts while maintaining its general language abilities.

\noindent\textbf{Instruction Tuning Stage:} The final stage focuses on developing instruction-following capabilities while maintaining strong video understanding. Using the Video-ChatGPT~\cite{maaz2023videochatgpt} instruction dataset (100k samples), we fine-tune the model with emphasis on generating appropriate responses to video-based instructions and questions.
\begin{table*}[!t]
    \centering
    \resizebox{\textwidth}{!}{
        \begin{tabular}{c|lcccccccccccc}
            \toprule
            \multicolumn{1}{c}{} & \multirow{2}{*}{\textbf{Model}} & \multirow{2}{*}{\textbf{Vision Size}} & \multirow{2}{*}{\textbf{Modality}} & \multirow{2}{*}{\textbf{Pretrain}}& \multirow{2}{*}{\textbf{Finetune}} & \multicolumn{2}{c}{\textbf{MSVD-QA}} & \multicolumn{2}{c}{\textbf{MSRVTT-QA}} & \multicolumn{2}{c}{\textbf{TGIF-QA*}} & \multicolumn{2}{c}{\textbf{Activity Net-QA}} \\
            \cmidrule{7-14}
            \multicolumn{1}{c}{} & & & & \textbf{Data} & \textbf{Data} & \textbf{Accuracy} & \textbf{Score} & \textbf{Accuracy} & \textbf{Score} & \textbf{Accuracy} & \textbf{Score} & \textbf{Accuracy} & \textbf{Score} \\
            \midrule
            & \multicolumn{13}{>{\columncolor{LightCyan}}c}{\textit{Encoder-based Vision-Language Models}} \\
            \multirow{9}{*}{\rotatebox[origin=c]{90}{Different Datasets}}
            & LLaMA Adapter \cite{zhang2024llamaadapter} & 404.3M & I & 567K & 52K & 54.9 & 3.1 & 43.8 & 2.7 & - & - & 34.2 & 2.7 \\
            & VideoChat \cite{li2023videochat} & 1.2B & V & 25M & 18K & 56.3 & 2.8 & 45.0 & 2.5 & 21.3 & 1.9 & 26.5 & 2.2 \\
            & Video-LLaMA \cite{zhang2023videollama}  & 1.1B & V & 3.1M & 164K & 51.6 & 2.5 & 29.6 & 1.8 & - & - &   12.4 & 1.1 \\
            & ChatUniVi \cite{jin2024chatunivi} & 307M & V+I & 1.6M & 649K & 65.0 & 3.6 & 54.6 & 3.1 & 38.2 & 3.0 & 45.8 & 3.2 \\
            & LLaMA-VID \cite{li2025llamavid} & 1B & V+I & 790K & 763K & 69.7 & 3.7 & 57.7 & 3.2 & - & - & 47.4 & 3.3 \\
            & Video-LLaVA \cite{lin2023videollava}  & 425M & V+I & 1.26M & 765K & 70.7 & 3.9 & 59.2 & 3.5 & 47.0 & 3.3 & 45.3 & 3.3 \\
            & VideoChat2 \cite{li2024mvbench} & 496M & V+I & 37M & 2M & 70.0 & 3.9 & 54.1 & 3.3 & - & - & 49.1 & 3.3 \\
            & ST-LLM \cite{liu2024stllm} & 1.3B & V & - & 400K & 74.6 & 3.9 & 63.2 & 3.4 & - & - & 50.9 & 3.3 \\
            & \multicolumn{13}{>{\columncolor{LightCyan}}c}{\textit{Encoder-free Vision-Language Models}} \\
            & EVE \cite{diao2024eve} & 30M & I & 33M & 665K & 14.3 & 1.7  & - & - & - & - & 6.8 & 1.8 \\
            \midrule
            \multirow{6}{*}{\rotatebox[origin=c]{90}{Same Dataset}}
            & \multicolumn{13}{>{\columncolor{LightCyan}}c}{\textit{Encoder-based Vision-Language Models}} \\
            & Video-ChatGPT \cite{maaz2023videochatgpt}  & 307M & V & - & 100K & 64.9 & 3.3 & 49.3 & 2.8 & 40.7 & 3.1 & 35.2 & 2.8 \\
            & Video-LLaVA \cite{lin2023videollava} & 425M & V & 702K & 100K & 64.8 & - & 58.3 & - & 41.7 & - & 40.7 & - \\
            & \multicolumn{13}{>{\columncolor{LightCyan}}c}{\textit{Encoder-free Vision-Language Models}} \\
            & EVE*  & 30M & V & 702K & 100K & 60.5 & 3.3 & 49.7 & 3.0 & 39.2 & 2.9 & 38.1 & 3.0 \\
            & Video-Panda (ours) & 45M & V & 702K & 100K & 64.7 & 3.8 & 54.8 & 3.4 & 42.9 & 3.2 & 40.0 & 3.3 \\
            \bottomrule
        \end{tabular}
    }
    \caption{Comparison with other video-language models that use LLMs with 7B parameters on open-ended video question answering. \textit{Vision Size} refers to \#parameters of vision encoder and alignment modules. \textit{Modality} indicates whether videos and/or images are used as training data. For \textit{TGIF-QA*}, we re-evaluated the results since the performance depends on the current version of GPT-3.5 which changes over time and highly impacts the evaluation. \textit{EVE*} is our extension of EVE  \cite{diao2024eve} to video data. 
    }
    \label{tab:video_sota}
\end{table*}

\section{Experiments}
\label{sec:experiments}

\subsection{Datasets and Implementation Details}

\noindent\textbf{Training Data:} Our training dataset is organized into three distinct stages. The first stage uses a randomly sampled subset of 351K video-text pairs (50\%) from Valley~\cite{luo2023valley}, which contains 702K video-text pairs sourced from the WebVid~\cite{bain2021frozen} dataset. Stage two expands to the full Valley dataset of 702K pairs. The final stage incorporates an additional 100K video-text instruction dataset from Video-ChatGPT~\cite{maaz2023videochatgpt}, focusing on complex visual reasoning tasks.

\noindent\textbf{Open-Ended Video QA Evaluation Data:} We evaluate our approach on four open-ended VideoQA datasets: MSRVTT-QA~\cite{xu2016msrvtt}, MSVD-QA~\cite{chen2011msvd}, TGIF-QA~\cite{jang2017tgif}, and ActivityNet-QA~\cite{yu2019activitynet}. MSRVTT-QA and MSVD-QA include five question types (what, who, how, when, and where), while TGIF-QA uses four types (object, number, color, and location). MSRVTT-QA contains 10K video clips and 243K question-answer pairs (158K/12K/73K for training/validation/testing), while MSVD-QA includes 1,970 video clips and 51K question-answer pairs (32K/6K/13K split). TGIF-QA comprises 46K GIFs and 53K question-answer pairs (39K/13K for training/testing). ActivityNet-QA, featuring longer videos averaging three minutes, covers nine question types (motion, spatial, temporal, yes-no, color, object, location, number, and other) with 5,800 videos and 58K question-answer pairs (32K/18K/8K split). We implemented a zero-shot evaluation protocol utilizing GPT-assisted assessment to analyze the model's performance. The evaluation methodology uses both prediction accuracy and a 5-point scale assessment.

\noindent\textbf{Fine-Grained Video QA Evaluation Data:} For a more fine-grained evaluation and analysis of our method, we adopt the benchmark introduced by Video-ChatGPT~\cite{maaz2023videochatgpt}. Built upon the ActivityNet-200~\cite{caba2015activitynet} dataset, this framework assesses responses across five dimensions: correctness of information, detail orientation, contextual understanding, temporal understanding, and consistency. Each aspect is scored on a scale of 1-5, evaluating factual accuracy, response completeness, context alignment, sequential comprehension, and reliability across similar queries.

\begin{table}[!t]
    \centering
    \resizebox{\columnwidth}{!}{
        \begin{tabular}{c|lccccccc}
        \toprule
        \multicolumn{1}{c}{} & \textbf{Model} & \textbf{Correctness} & \textbf{Detail} & \textbf{Context} & \textbf{Temporal} & \textbf{Consistency} & \textbf{AVG} \\
        \midrule
        & \multicolumn{7}{>{\columncolor{LightCyan}}c}{\textit{Encoder-based Vision-Language Models}} \\
        \multirow{7}{*}{\rotatebox[origin=c]{90}{Different Datasets}}
        & VideoChat \cite{li2023videochat} & 2.23 & 2.50 & 2.53 & 1.94 & 2.24 & 2.29 \\
        & LLaMA Adapter \cite{zhang2024llamaadapter} & 2.03 & 2.32 & 2.30 & 1.98 & 2.15 & 2.16\\
        & Video-LLaMA \cite{zhang2023videollama} & 1.96 & 2.18 & 2.16 & 1.82 & 1.79 & 1.98 \\
        & ChatUniVi \cite{jin2024chatunivi} & 2.89 & 2.91 & 3.46 & 2.39 & 2.81 & 2.89 \\
        & LLaMA-VID \cite{li2025llamavid} & 2.96 & 3.00 & 3.53 & 2.46 & 2.51& 2.89 \\
        & Video-LLaVA \cite{lin2023videollava} & 2.84 & 2.86 & 3.44 & 2.46 & 2.57 & 2.81 \\
        & VideoChat2 \cite{li2024mvbench} & 3.02 & 2.88 & 3.51 & 2.66 & 2.81 & 2.98 \\
        & ST-LLM \cite{liu2024stllm} & 3.23 & 3.05 & 3.74 & 2.93 & 2.81 & 3.15 \\
        \midrule
        \multirow{5}{*}{\rotatebox[origin=c]{90}{Same Datasets}}
        & \multicolumn{7}{>{\columncolor{LightCyan}}c}{\textit{Encoder-based Vision-Language Models}} \\
        & Video-ChatGPT \cite{maaz2023videochatgpt} & 2.40 & 2.52 & 2.62 & 1.98 & 2.37 & 2.38 \\
        & Video-LLaVA* \cite{lin2023videollava} & 2.46 & 2.37 & 2.89 & 2.12 & 2.17 & 2.40 \\
        & \multicolumn{7}{>{\columncolor{LightCyan}}c}{\textit{Encoder-free Vision-Language Models}} \\
        & Video-Panda (ours) & 2.74 & 2.47 & 3.01 & 2.26 & 2.36 & 2.57  \\
        \bottomrule
        \end{tabular}
    }
    \caption{Comparison of video-language models on fine-grained video question answering metrics (scale 1-5) across correctness, detail, context, temporal reasoning, and consistency. \textit{Video-LLaVA*}: trained with video-only datasets for fair comparison.}
    \label{tab:video_sota2}
    \vspace{-0.1cm}
\end{table}

\noindent\textbf{Implementation Details:}
Our model is built on Vicuna-7B v1.5~\cite{vicuna2023} tailored to vision-language applications. The teacher model is LanguageBind \cite{zhu2023languagebind}, which is initialized from ViT-L/14~\cite{alexey2020vit} and pre-trained with contrastive objectives to align multi-modal representations within a unified embedding space. Eight frames are sampled uniformly from each video, and the maximum image/frame width or height is limited to 448 pixels. Input to the teacher model is padded and resized to 224x224. Training samples for the three stages are 351K, 702K, and 100K, respectively. For each training batch, cross-attentions are performed between query and valid spatial tokens by removing padded tokens for efficient modeling. We keep other hyperparameters the same as previous work~\cite{diao2024eve}. The full hyperparameter settings can be found in the supplementary material.

\begin{table}[!t]
\hfill
    \centering
    \resizebox{1.0\linewidth}{!}{
        \begin{tabular}{lccc}
            \toprule
            \textbf{Model} & \textbf{Vision Size (M)} & \textbf{FLOPs (G)} & \textbf{Latency (ms)}\\
            \midrule
            Video-ChatGPT \cite{maaz2023videochatgpt} & 307 & 8109.1 & 171\\
            Video-LLaVA \cite{lin2023videollava} & 425 & 864.1 & 125 \\
            \midrule
            Video-Panda & 45 & 105.5 & 41\\
            \bottomrule
        \end{tabular}
    }
    \caption{
    Comparison of the number of parameters, FLOPs, and latency of the vision part.
    }
    \label{tab:inference}
\end{table}

\subsection{Quantitative Evaluation}
In this subsection, we present a comprehensive evaluation of our model's performance. We analyze the open-ended and fine-grained video understanding capabilities across multiple benchmark datasets, comparing our encoder-free approach with approaches trained on the same dataset as well as approaches that use more or other data for training. Subsequently, we examine the computational efficiency through parameter count and inference time.

\noindent\textbf{Open-Ended Video Question Answering:} The results of our evaluation for open-ended video question answering are presented in \autoref{tab:video_sota}. Our proposed Video-Panda model, despite being the only encoder-free video-language model, demonstrates competitive performance with encoder-based approaches when compared to methods that have been trained on the same dataset (bottom). We first compare the results of Video-Panda with Video-ChatGPT \cite{maaz2023videochatgpt}. While Video-ChatGPT achieves a slightly higher accuracy on MSVD-QA (64.7\% vs 64.9\%), Video-Panda achieves a higher score of 3.8. For the other datasets MSRVTT-QA, TGIF-QA, and Activity Net-QA, Video-Panda outperforms Video-ChatGPT for all metrics. In comparison to Video-LLaVA \cite{lin2023videollava}, which has even more parameters than Video-ChatGPT, Video-Panda achieves competitive results and even outperforms it on TGIF-QA. Since EVE~\cite{diao2024eve} performs poorly on the video benchmarks, we also extended EVE to the video domain (see supplementary material). Extending EVE to the video domain, however, performs substantially worse than Video-Panda. This shows the importance of the proposed video-specific alignment module. For completeness, we also compare to methods that have been pre-trained and fine-tuned on other datasets although the results are not directly comparable. Nevertheless, Video-Panda also outperforms VideoChat \cite{li2023videochat} and Video-LLaMA \cite{zhang2023videollama}, which are trained only on video as well. 

\begin{table}[t]
    \centering
    \resizebox{.45\textwidth}{!}{
        \begin{tabular}{lcc}
            \toprule
            \textbf{Model} & \textbf{MSVD-QA} & \textbf{Activity Net-QA} \\
            \midrule
            \rowcolor{LightCyan} \multicolumn{3}{c}{\textit{Spatial}} \\
            w/o $\texttt{<row>}$ & 63.2/3.7 & 39.5/3.3 \\
            w/o FSRA & 63.4/3.7 & 39.2/3.3 \\ 
            w/o LSD (avg pool) & 58.0/3.6 & 38.1/3.2 \\
            \midrule
            \rowcolor{LightCyan} \multicolumn{3}{c}{\textit{Temporal}} \\
            w/o LSTE & 63.6/3.7 & 39.4/3.3 \\
            w/o GSTRA & 63.0/3.7 & 38.2/3.2 \\
            w/o GSTRA \& LSTE & 62.2/3.7 & 38.1/3.2 \\
            \midrule
            Video-Panda & \textbf{64.7/3.8} & \textbf{40.0/3.3} \\
            \bottomrule
        \end{tabular}
    }
    \caption{Ablation study on the impact of removing different spatial and temporal modules used in our design.}
    \label{tab:ablation_design}
\end{table}
\begin{table}[t]
   \centering
   \resizebox{1.0\linewidth}{!}{
       \begin{tabular}{lcc}
           \toprule
           \textbf{Distillation Loss} & \textbf{MSVD-QA} & \textbf{Activity Net-QA} \\
           \midrule
           w/o Distillation & 63.1/3.7 & 39.8/3.3  \\
           Mean Squared Error & 63.5/3.7 & 38.2/3.2 \\
           Negative Cosine Similarity & \textbf{64.7/3.8} & \textbf{40.0/3.3} \\
           \bottomrule
       \end{tabular}
   }
   \caption{Ablation study on the impact of not using distillation loss or a different (MSE) loss.}
   \label{tab:ablation_loss}
   \vspace{-0.1cm}
\end{table}

\noindent\textbf{Fine-Grained Video Question Answering:} The results for fine-grained video question answering are presented in \autoref{tab:video_sota2}. Our proposed Video-Panda model, despite being an encoder-free approach with only 45M parameters as shown in \autoref{tab:video_sota}, demonstrates strong performance when compared to methods trained on the same dataset. Video-Panda outperforms Video-LLaVA in all aspects. When compared to Video-ChatGPT, Video-Panda shows substantial improvements in correctness (2.74 vs 2.40), context (3.01 vs 2.62), and temporal understanding (2.26 vs 1.98), although it achieves slightly lower scores in detail and consistency. It is worth noting that Video-ChatGPT processes 100 frames per video while Video-Panda uses only 8 frames, making our model's performance particularly impressive. We also compare to approaches that have been pre-trained and fine-tuned on different datasets.

\begin{figure*}[t]
  \centering
    \includegraphics[width=\linewidth]{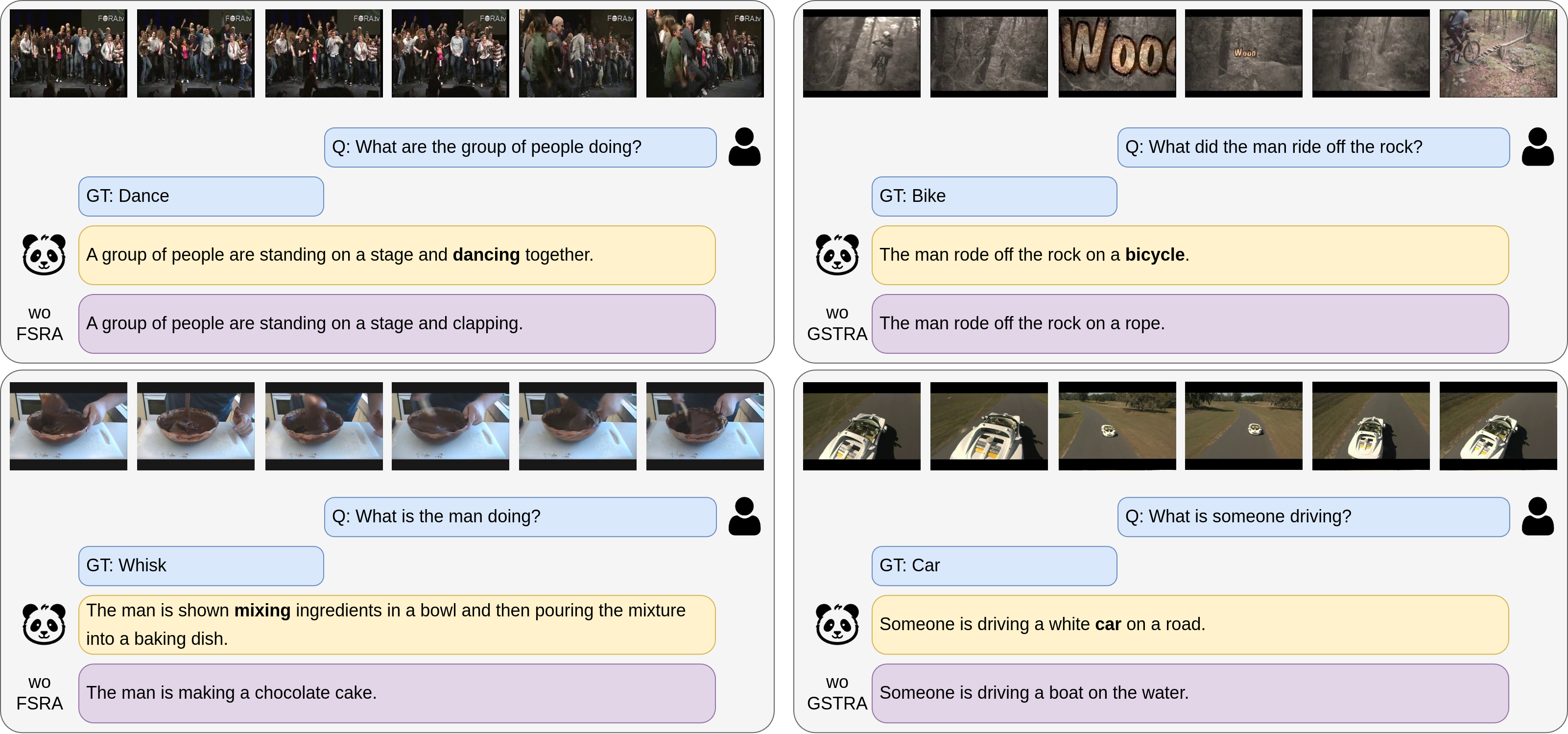} 
    \caption{Qualitative examples showing the impact of removing Frame-wise Spatial Relationship Aggregator (FSRA) and Global Spatio-Temporal Relationship Aggregator (GSTRA).}
   \label{fig:qualitative}
\end{figure*}

\noindent\textbf{Parameters and Inference Time:} 
As shown in \autoref{tab:inference}, our Video-Panda model requires much fewer parameters (45M) for its visual component compared to VideoChatGPT (307M) and Video-LLaVA (425M). This also translates directly to better inference speed, which is very important for practical applications. Video-Panda processes videos in 41ms, approximately 4$\times$ faster than VideoChatGPT (171ms) and 3$\times$ faster than Video-LLaVA (125ms). These results demonstrate that our encoder-free approach not only maintains competitive performance but also offers significant computational advantages.
\section{Analysis and Ablations}

To evaluate our design choices and their impact on model performance, we conduct ablation studies regarding architectural components, distillation loss, and qualitative impact of the spatial and temporal aggregators. Additional ablations are presented in the supplementary materials.

\noindent\textbf{Component Analysis:} We first analyze the contribution of each architectural component in \autoref{tab:ablation_design}. The removal of row tokens ($\texttt{<row>}$) leads to a noticeable drop in performance, particularly on MSVD-QA (1.5\% decrease), highlighting the importance of maintaining spatial structure information. Similarly, removing the Frame-wise Spatial Relationship Aggregator (FSRA) results in a decrease of 1.3\%. Replacing Local Spatial Downsampling (LSD) by simply average pooling decreases the performance on MSVD-QA by 6.7\%. 
The Local Spatio-Temporal Encoding (LSTE) and Global Spatio-Temporal Relationship Aggregator (GSTRA) modules show complementary benefits, with their combined removal resulting in a 2.5\% drop on MSVD-QA, indicating the effectiveness of our spatio-temporal modeling approach.

\noindent\textbf{Distillation Loss:} The choice of the distillation loss function shows some impact on the model performance, as shown in \autoref{tab:ablation_loss}. Our negative cosine similarity loss outperforms MSE by 1.2\% on MSVD-QA and 1.8\% on ActivityNet-QA. Removing distillation entirely still maintains relatively strong performance, particularly on ActivityNet-QA. 
This suggests that while our distillation approach provides meaningful benefits, the architectural design itself captures significant video understanding capabilities. The superior performance of cosine similarity over MSE indicates the importance of direction-based alignment rather than exact magnitude matching in feature space.

\noindent\textbf{Qualitative Ablations:} The qualitative examples in \autoref{fig:qualitative} demonstrate the critical role of our FSRA and GSTRA components in accurate visual understanding. Without FSRA, the model struggles to properly interpret more granular actions, as evidenced by misinterpreting a coordinated dance performance as mere ``clapping'' and failing to recognize the ``mixing'' action in favor of the end goal of ``making a chocolate cake''. When removing GSTRA, the model tends to focus on single frames instead of considering the context of the entire video, which can be misleading. For instance, the construction at the end of the video is misinterpreted as ``rope'' and the ``car'' on a ``road'' is misinterpreted as a ``boat'' on the ``water'' at the beginning. This shows that the global video context from GSTRA as well as the frame-wise representation from FSRA are needed.
\section{Conclusion}

This work introduces an efficient encoder-free approach to video-language understanding through our Spatio-Temporal Alignment Block (STAB). We reduce the number of visual processing parameters by 6.5$\times$ or greater while maintaining competitive performance or outperforming comparable approaches in open-ended video question answering, obtaining 64.7\% accuracy on MSVD-QA and 54.8\% on MSRVTT-QA, as well as fine-grained video understanding. Additionally, STAB processes videos 3-4$\times$ faster than encoder-based methods. Our method shows promise for making video-language models more efficient, potentially benefiting longer videos and other tasks.

\clearpage
\subsection*{Acknowledgement}
This work was supported by the Federal Ministry of Education and Research (BMBF) under grant no.\ 01IS22094A WEST-AI and the ERC Consolidator Grant FORHUE (101044724). Yanan Luo has been supported by the Chinese Scholarship Council (202108440041).
For the computations involved in this research, we acknowledge EuroHPC Joint Undertaking for awarding us access to Leonardo at CINECA, Italy, through EuroHPC Regular Access Call - proposal No.\ EHPC-REG-2024R01-076. 

{
    \small
    \bibliographystyle{ieeenat_fullname}
    \bibliography{main}
}

\clearpage
\appendix

\begin{center}
\textbf{{\Large Appendix}}
\end{center}

\noindent This section contains supplemental material, offering further results and analysis to complement the main paper. We provide additional details on the following topics:
\begin{itemize}
    \item Detailed Hyperparameters (\autoref{sec:app:hparams})
    \item Additional Ablations (\autoref{sec:app:ablations})
    \item Additional Dataset Details (\autoref{sec:app:dataset})
    \item{Evaluation on Long Videos}
    (\autoref{sec:app:eval_long_video})
    \item Additional Qualitative Ablations (\autoref{sec:app:quali})
    \item EVE Baseline for Videos (\autoref{sec:app:eve})
    \item Broader Impact (\autoref{sec:app:bimpact})
\end{itemize}

\section{Detailed Hyperparameters}
\label{sec:app:hparams}

In \autoref{tab:hyperparam} we provide comprehensive hyperparameter configurations for Video-Panda's three-stage training process.

\begin{table}[ht!] 
    \centering
    \resizebox{\columnwidth}{!}{
        \begin{tabular}{lccc}
            \toprule
             Hyperparameter & Stage-1 & Stage-2 & Stage-3 \\
             \midrule
             Batch Size & 2048 & 2048 & 1024 \\
             Learning Rate (lr) & 4e-4 & 4e-5 & 2e-5 \\
             LR Schedule & cos. decay  & cos. decay & cos. decay \\ 
             LR Warmup Ratio & 0.03 & 0.01 & 0.01 \\
             Weight Decay & 0 & 0 & 0\\
             Epoch & 1 & 1 & 1 \\
             Optimizer & AdamW & AdamW & AdamW \\
             DeepSpeed Stage & 2 & 2 & 2 \\
            LLM & Frozen & Trainable & Trainable \\
            STAB & Trainable & Trainable & Trainable \\
            
             \bottomrule
        \end{tabular}
    }
    \caption{Hyperparameter Settings}
    \label{tab:hyperparam}
    \vspace{-0.1cm}
\end{table}

\section{Additional Ablations}
\label{sec:app:ablations}

\noindent\textbf{Training Data for Initial Alignment:} \autoref{tab:ablation_data} shows the impact of data scale during initial alignment. Using the full dataset (702K samples) in Stage 1 yields marginally lower performance compared to using half (351K samples). This suggests our staged training approach benefits from gradual complexity scaling, allowing the model to establish robust representations before incorporating the complete dataset in later stages.

\begin{table}[t]
    \centering
    \resizebox{\columnwidth}{!}{
        \begin{tabular}{lcc}
            \toprule
            \textbf{\#Samples for Initial Alignment} & \textbf{MSVD-QA} & \textbf{Activity Net-QA} \\
            \midrule
            702K Video-Text Pairs (full) & 63.7/3.8 & 39.7/3.3 \\
            351K Video-Text Pairs (half) & \textbf{64.7/3.8} & \textbf{40.0/3.3} \\
            \bottomrule
        \end{tabular}
    }
    \caption{Ablation study on amount of data for the first training stage.}
    \label{tab:ablation_data}
\end{table}

\noindent\textbf{Downsampling Position:} Regarding temporal layer placement in \autoref{tab:ablation_downsampling_pos}, we find that applying LSD after LSTE improves performance on all datasets except of ActivityNet-QA. 
For consistency across our experiments, we maintain LSD placement after LSTE.

\begin{table}[ht]
   \centering
   \resizebox{\columnwidth}{!}{
       \begin{tabular}{lcccc}
           \toprule
           \textbf{Model} & \textbf{MSVD-QA} & \textbf{MSRVTT-QA} & \textbf{TGIF-QA} & \textbf{Activity Net-QA} \\
           \midrule
           Before LSTE & 64.2/3.8 & 54.6/3.4 & 42.7/3.2 & \textbf{42.3/3.3} \\
           After LSTE (Ours) & \textbf{64.7/3.8} & \textbf{54.8/3.4} & \textbf{42.9/3.2} & 40.0/3.3 \\
           \bottomrule
       \end{tabular}
   }
   \caption{Ablation study on downsampling positions of LSD.}
   \label{tab:ablation_downsampling_pos}
\end{table}

\noindent\textbf{Downsampling Strategy:} As shown in \autoref{tab:ablation_downsampling}, our LSD method outperforms alternative approaches. The Perceiver Resampler (PR) performs notably poorly (21.3\% lower on MSVD-QA), likely due to excessive information compression. While average pooling performs better, it still underperforms LSD by 6.7\%, demonstrating the superiority of our learnable downsampling approach.

\begin{table}[ht]
   \centering
   \resizebox{\columnwidth}{!}{
       \begin{tabular}{lcc}
           \toprule
           \textbf{Model} & \textbf{MSVD-QA} & \textbf{Activity Net-QA} \\
           \midrule
           w/o LSD (half-resolution) & 48.2/3.3 & 38.5/3.2 \\
           w/o LSD (avg pool) & 58.0/3.6 & 38.1/3.2 \\
           w/o LSD (PR) & 43.4/3.2 & 27.8/2.9 \\
          \midrule
           Video-Panda (LSD) & \textbf{64.7/3.8} & \textbf{40.0/3.3} \\
           \bottomrule
       \end{tabular}
   }
   \caption{Ablation study on downsampling methods. PR stands for Perceiver Resampler~\cite{alayrac2022flamingo}.}
   \label{tab:ablation_downsampling}
\end{table}
\begin{table}[t!]
\hfill
   \centering
   \resizebox{\columnwidth}{!}{
       \begin{tabular}{lcc}
           \toprule
           \textbf{Model} & \textbf{MSVD-QA} & \textbf{Activity Net-QA} \\
           \midrule
           CLIP & 60.3/3.5 & 38.6/3.2 \\
           InternVideov2 & 62.5/3.6 & 39.6/3.2 \\
           DINOv2 & 61.7/3.5 & 38.1/3.2\\
          \midrule
           LanguageBind (Video-Panda) & \textbf{64.7/3.8} & \textbf{40.0/3.3} \\
           \bottomrule
       \end{tabular}
   }
   \caption{Ablation study on different teacher encoders.}
   \setlength{\abovecaptionskip}{-10pt}
\setlength{\belowcaptionskip}{-10pt}
   \label{tab:ablation_teacher}
\end{table}
\vspace{-10cm}

\noindent\textbf{Different Teachers:} As shown in \autoref{tab:ablation_teacher}, LanguageBind consistently outperforms other teacher encoders across both datasets. While InternVideo achieves the second-best performance, it still falls short by 2.2\% on MSVD-QA and 0.4\% on Activity Net-QA. CLIP and DINOv2 show comparable performance to each other but lag behind LanguageBind by 3-4\%, demonstrating the effectiveness of our chosen teacher encoder.

\section{Additional Dataset Details}
\label{sec:app:dataset}

\noindent\textbf{Pre-training Dataset:} The Valley-Pretrain-702K dataset is a large-scale pre-training dataset designed for video-language understanding tasks. It comprises 702K video-text pairs from the WebVid dataset~\cite{bain2021frozen}, filtered by~\cite{luo2023valley} using methods established by LLaVA~\cite{liu2023llava} to optimize the balance between conceptual diversity and training efficiency. The dataset is structured as single-round dialogues, where each video is paired with questions about its content and corresponding caption-based answers.

\noindent\textbf{Fine-tuning Dataset:} The Video-ChatGPT-100K dataset was developed for fine-tuning video-language models, comprising 100K video instruction samples collected by~\cite{maaz2023videochatgpt}. The dataset combines human expertise with semi-automated methods to balance quality and scalability. Expert annotators provide detailed, context-rich descriptions that enhance the model's comprehension of complex video content. A semi-automatic framework leverages state-of-the-art vision-language models to generate large-scale annotations efficiently, ensuring substantial data volume while maintaining rigorous quality standards.

\noindent\textbf{Fine-Grained Video QA Evaluation Dataset: } We evaluate fine-grained video question answering using the Video-based Text Generation Performance Benchmarking methodology developed by Video-ChatGPT~\cite{maaz2023videochatgpt}. This benchmark provides a comprehensive evaluation framework for assessing text generation in video-based conversational models. Using the ActivityNet-200 dataset~\cite{caba2015activitynet}, which contains videos with descriptive captions and human-annotated question-answer pairs, the framework implements a systematic evaluation approach. The methodology utilizes GPT-3.5 to evaluate models across multiple dimensions on a scale of 1 to 5. The assessment criteria include:

\begin{enumerate}[label=(\roman*)]
    \item \textit{Correctness of Information:} Evaluates accuracy of generated text and its alignment with video content.
    \item \textit{Detail Orientation:} Assesses response comprehensiveness, examining both coverage of major points and specificity of details.
    \item \textit{Contextual Understanding:} Measures the model's ability to interpret and respond within the video's broader context.
    \item \textit{Temporal Understanding:} Evaluates the model's capacity to track and articulate the chronological sequence of events.
    \item \textit{Consistency:} Assesses the model's ability to maintain coherent responses across different questions and video segments.
\end{enumerate}

\section{Evaluation on Long Videos}
\label{sec:app:eval_long_video}

To explore the potential of Video-Panda on long video benchmarks, we have evaluated our method on the EgoSchema~\cite{mangalam2023egoschema} and Video-MME-M~\cite{fu2024videomme} datasets. The results presented in~\autoref{tab:other_benchmarks} confirm the results presented in the paper, i.e., we achieve similar or slightly better accuracy compared to Video-ChatGPT and Video-LLaVA, but require much less computational resources (\autoref{tab:inference}). 

\begin{table}[!t]
\hfill
    \centering
    \resizebox{\columnwidth}{!}{
        \begin{tabular}{lccccc}
            \toprule
            \textbf{Model} & \textbf{Vision Size (M)} & \textbf{EgoSchema} & \textbf{VideoMME-M} \\ 
            \midrule
            Video-ChatGPT & 307 & 34.2 & 36.0 \\ 
            Video-LLaVA & 425 & \underline{36.1} & \textbf{38.1 }\\
            \midrule
            Video-Panda & 45 & \textbf{36.4} & \underline{37.9}\\
            \bottomrule
        \end{tabular}
}
\caption{Results on the EgoSchema and VideoMME-M datasets. }
\label{tab:other_benchmarks}
\vspace{-0.5cm}
\end{table}

\section{Additional Qualitative Ablations}
\label{sec:app:quali}
\begin{figure*}[t]
  \centering
    \includegraphics[width=\linewidth]{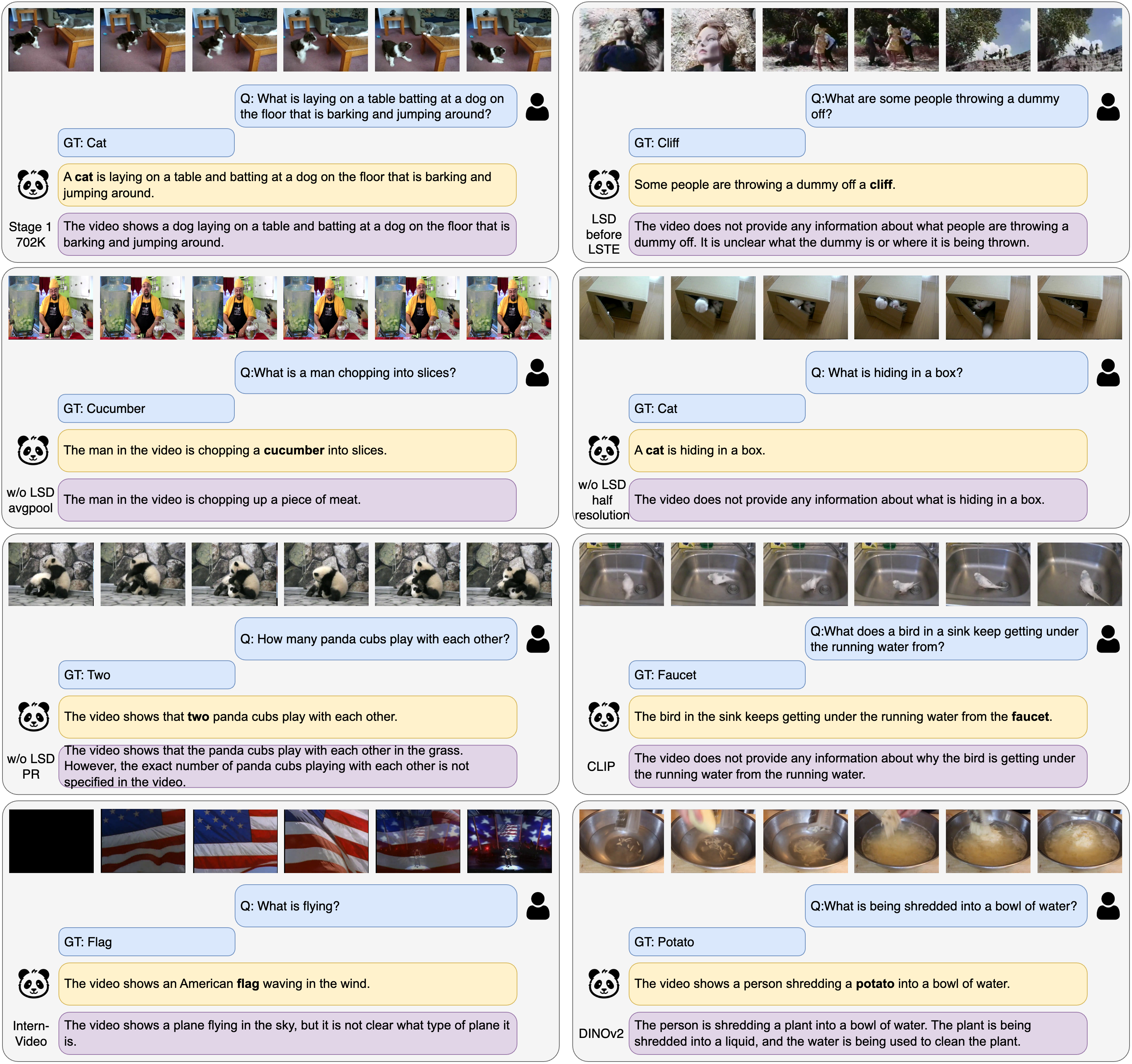} 
   \caption{\textbf{Qualitative comparisons of different design choices of Video-Panda:} The figure presents eight video examples with ground truth (GT) annotations and model predictions under different training configurations. The \textit{top row} demonstrates the effect of 702K training samples in stage 1 (\textit{left}) and the impact of performing Local Spatial Downsampling (LSD) before Local Spatial-Temporal Encoding (LSTE) (\textit{right}). The \textit{second row} shows results from removing LSD while using average pooling (\textit{left}), half-resolution (\textit{right}), and perceiver resampler (\textit{third row left}). The \textit{third row right} and \textit{fourth row} illustrate the effects of different teacher models for knowledge distillation: CLIP (\textit{third row right}), Intern-Video (\textit{left}), and DINOv2 (\textit{right}). Each example includes the original model prediction (yellow) and an ablated version (purple), highlighting how architectural and training choices affect Video-Panda's ability to interpret dynamic visual scenes and answer questions.
   The qualitative examples are from the MSVD-QA dataset. 
   }
   \label{fig:appendix_quali}
\end{figure*}

\begin{figure*}[ht]
  \centering
    \includegraphics[width=\linewidth]{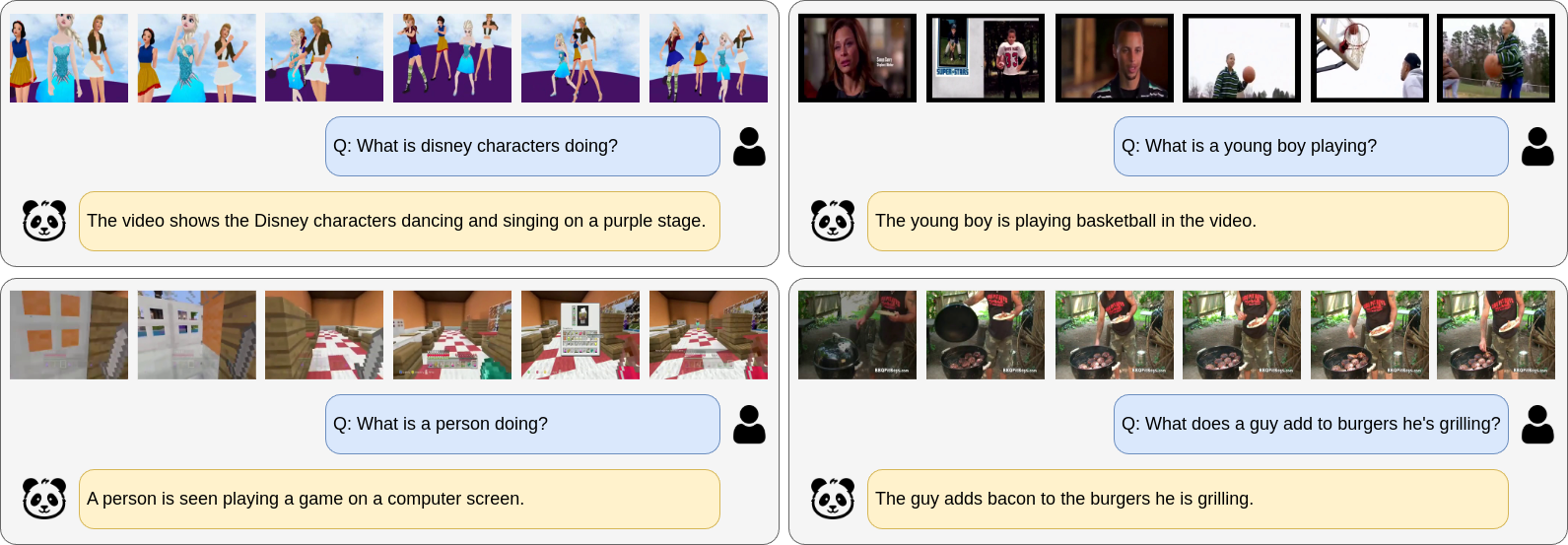} 
    \caption{Qualitative examples from the MSRVTT-QA dataset.}
   \label{fig:demo_msrvtt}
\end{figure*}

\begin{figure*}[ht]
  \centering
    \includegraphics[width=\linewidth]{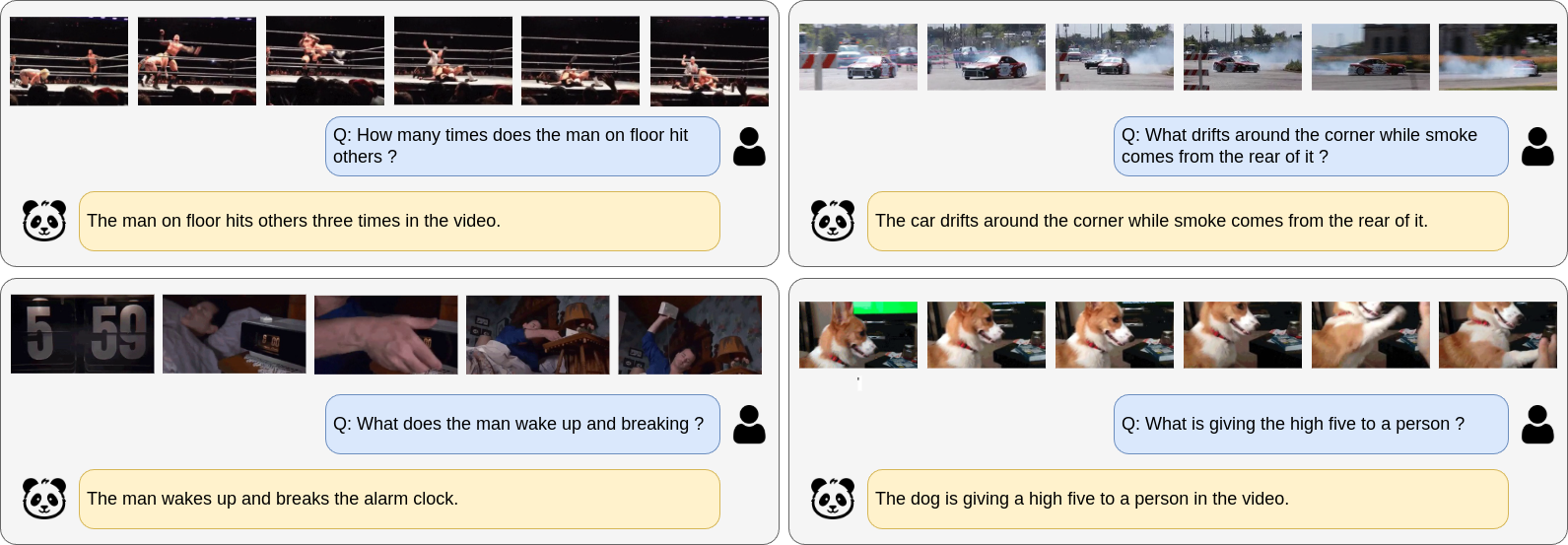} 
    \caption{Qualitative examples from the TGIF-QA dataset.}
   \label{fig:demo_tgif}
\end{figure*}

\begin{figure*}[ht]
  \centering
    \includegraphics[width=\linewidth]{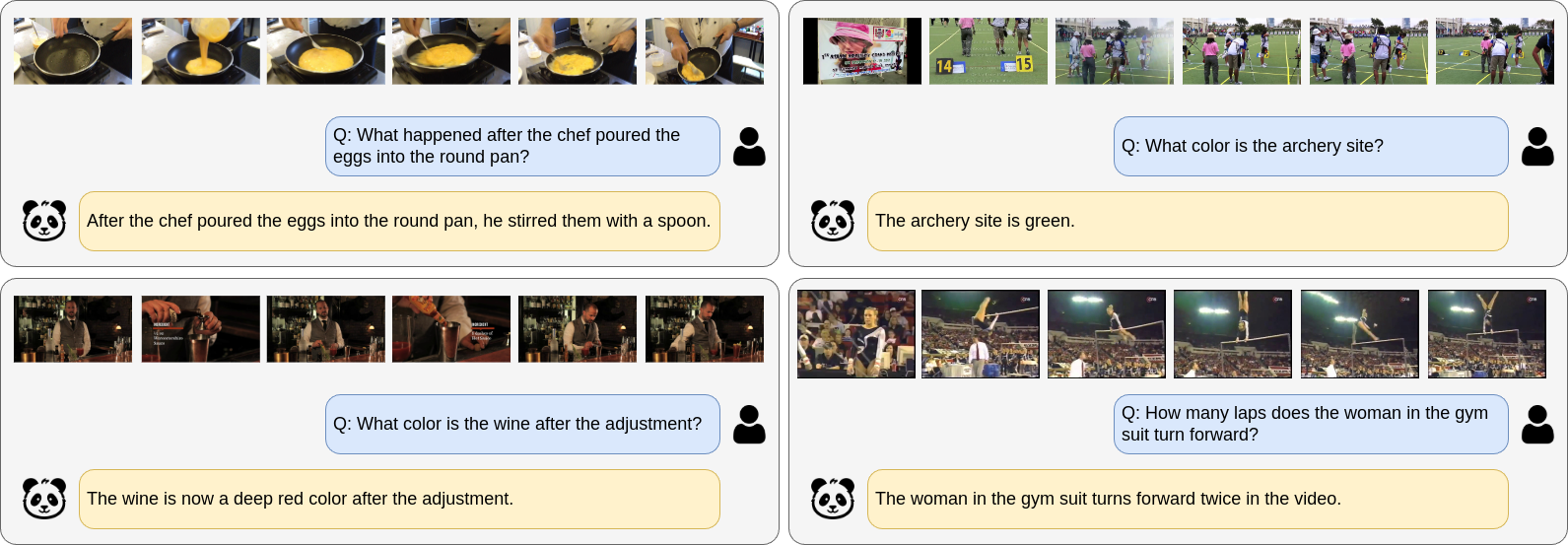} 
    \caption{Qualitative examples from the ActivityNet-QA dataset.}
   \label{fig:demo_actnet}
\end{figure*}

We present additional qualitative examples of our ablation studies in \autoref{fig:appendix_quali}, demonstrating Video-Panda's effectiveness across various video understanding tasks. When using the complete training dataset in Stage 1 (left-top example), the model exhibits overfitting tendencies due to data imbalance, as evidenced by the example showing dog interactions—likely influenced by the disparity between dog (7,807) and cat (5,050) instances in the Valley dataset.
The right-top example reveals that placing the LSD module before LSTE impairs cliff recognition due to early token downsampling and information loss. Models using alternative approaches (average pooling, half resolution, or perceiver resampler) struggle with content recognition (e.g., cucumber, cat, pandas) compared to our learnable downsampling approach. Additionally, models using image-based teachers (CLIP and DINOv2) tend to make frame-specific predictions rather than considering global context, as demonstrated by their failure to recognize shredded potatoes across multiple frames.
We also provide additional qualitative examples on each dataset in \autoref{fig:demo_msrvtt}, \autoref{fig:demo_tgif}, and \autoref{fig:demo_actnet}.

\section{EVE Baseline for Videos}
\label{sec:app:eve}

As the original EVE model~\cite{diao2024eve} was designed for image processing, we conducted a fair comparison by re-training it (denoted as \textit{EVE*} in \autoref{tab:video_sota}) using identical video data (Valley-702K and Video-ChatGPT-100K). For processing videos, each frame was treated independently as a separate image, with CLIP-ViT-L/14~\cite{radford2021clip} serving as the teacher model for distillation. While this approach enables frame-level analysis, it neglects temporal relationships. In our implementation, we employ Learnable Selective Downsampling (LSD) to process video frames efficiently, reducing each frame to a consistent token count while preserving essential information. The resulting tokens are flattened into a single sequence, with special split tokens inserted between frame representations to maintain frame boundaries and enable temporal relationship learning.

\section{Broader Impact}
\label{sec:app:bimpact}

We introduce Video-Panda, an encoder-free Video Language Model for video understanding. Our model addresses key practical challenges in large-scale AI deployment. While many VLMs raise concerns about data bias, privacy, and computational costs, Video-Panda mitigates these issues through two key design choices: training exclusively on publicly available datasets and eliminating the need for a pretrained encoder. This approach not only reduces ethical concerns but also significantly lowers computational requirements and deployment costs, making the model more accessible and environmentally sustainable.

\end{document}